\documentclass[10pt,twocolumn,letterpaper]{article}
\usepackage{enumitem}
\usepackage{iccv}
\usepackage{times}
\usepackage{epsfig}
\usepackage{graphicx}
\usepackage{amsmath}
\usepackage{amssymb}
\usepackage{multirow}
\usepackage{subcaption}
\usepackage{xspace}
\usepackage{bbding}
\usepackage{pifont}
\usepackage{color,xcolor}
\usepackage{stfloats}
\newcommand{\sgdg}{\textcolor{lightgray}{SGD}}
\usepackage{wasysym}
\newcommand{\cmark}{\ding{51}\xspace}%
\newcommand{\cmarkg}{\textcolor{lightgray}{\ding{51}}\xspace}%
\newcommand{\xmark}{\ding{55}\xspace}%
\newcommand{\xmarkg}{\textcolor{lightgray}{\ding{55}}\xspace}%

\newcommand{\blue}[1]{\textcolor{blue}{#1}}

\usepackage{caption}
\usepackage[pagebackref=true,breaklinks=true,letterpaper=true,colorlinks,bookmarks=false]{hyperref}
\iccvfinalcopy 

\def\iccvPaperID{3846} 

\ificcvfinal\fi

\begin{document}

\title{TransReID: Transformer-based Object Re-Identification}

\newcommand*{\affaddr}[1]{#1}
\newcommand*{\affmark}[1][*]{\textsuperscript{#1}}
\author{Shuting He\affmark[1,2]\thanks{This work was done when Shuting He was intern at Alibaba supervised by Hao Luo and Pichao Wang.}, Hao Luo\affmark[1], Pichao Wang\affmark[1], Fan Wang\affmark[1],  Hao Li\affmark[1], Wei Jiang\affmark[2]\\
\affaddr{\affmark[1]Alibaba Group},
\affaddr{\affmark[2]Zhejiang University} \\
{  \tt\small\{shuting\_he,jiangwei$\_$zju\}@zju.edu.cn }
{  \tt\small\{michuan.lh,pichao.wang,fan.w,lihao.lh\}@alibaba-inc.com }
}

\maketitle
\ificcvfinal\thispagestyle{empty}\fi

\begin{abstract}
Extracting robust feature representation is one of the key challenges in object re-identification (ReID). Although convolution neural network (CNN)-based methods have achieved great success, they only process one local neighborhood at a time and suffer from information loss on details caused by convolution and downsampling operators (\it{e.g.} pooling and strided convolution).
To overcome these limitations, we propose a pure transformer-based object ReID framework named TransReID. Specifically, we first encode an image as a sequence of patches and build a transformer-based strong baseline with a few critical improvements, which achieves competitive results on several ReID benchmarks with CNN-based methods.
To further enhance the robust feature learning in the context of transformers, two novel modules are carefully designed. (i) The jigsaw patch module (JPM) is proposed to rearrange the patch embeddings via shift and patch shuffle operations which generates robust features with improved discrimination ability and more diversified coverage. (ii) The side information embeddings (SIE) is introduced to mitigate feature bias towards camera/view variations by plugging in learnable embeddings to incorporate these non-visual clues.
To the best of our knowledge, this is the first work to adopt a pure transformer for ReID research. Experimental results of TransReID are superior promising, which achieve state-of-the-art performance on both person and vehicle ReID benchmarks. Code is available at \url{https://github.com/heshuting555/TransReID}.
\end{abstract}

\vspace{-1.5em}


\section{Introduction}

\begin{figure}[t]
    \centering
	\includegraphics[width=0.49\textwidth]{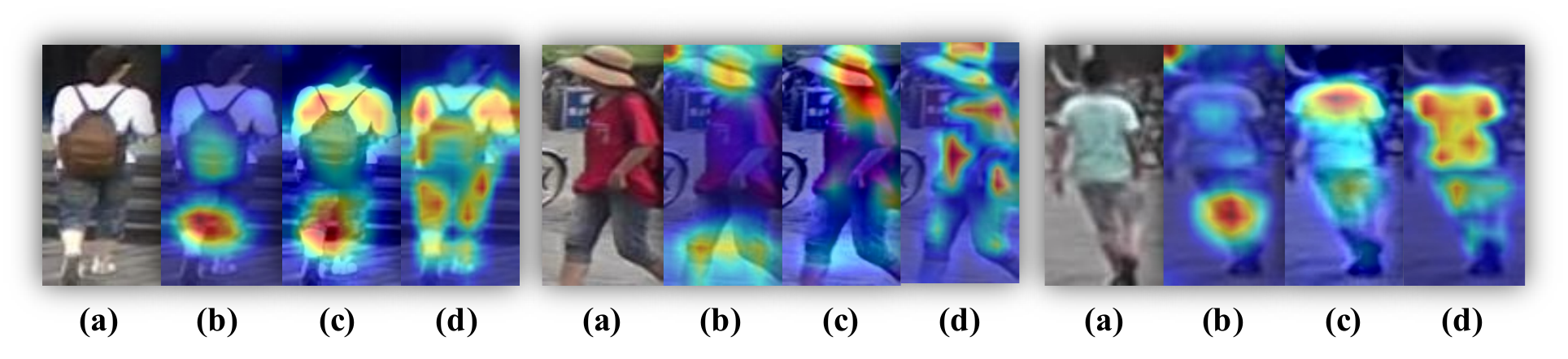}
	\vspace{-1.2em}
	\caption{Grad-CAM \cite{grad_cam} visualization of attention maps: (a) Original images, (b) CNN-based methods, (c) CNN+attention methods, (d) Transformer-based methods which captures global context information and more discriminative parts.}
	\vspace{-1.0em}
	\label{fig:motivation}
\end{figure}
\begin{figure}[t]
    \centering
	\includegraphics[width=0.49\textwidth]{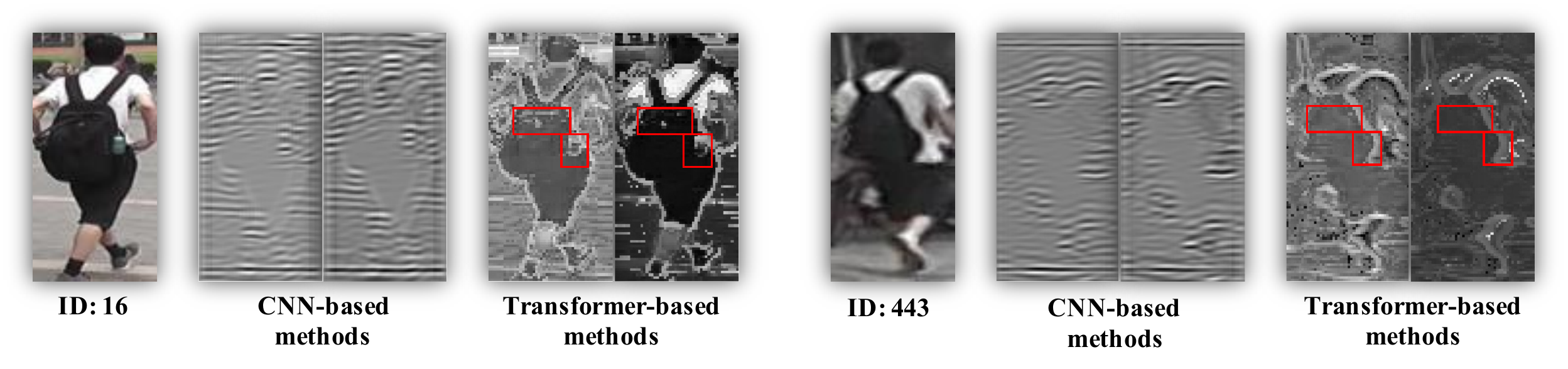}
	\vspace{-1.8em}
	\caption{Visualization of output feature maps for 2 hard samples with similar appearances. Transformer-based methods retain backpack details on output feature maps in contrast to CNN-based methods, as noted in red boxes. For better visualization, input images are scaled to size $1024\times512$.}
	\vspace{-1.em}
	\label{fig:feature_map}
\end{figure}

Object re-identification (ReID) aims to associate a particular object across different scenes and camera views, such as in the applications of person ReID and vehicle ReID. Extracting robust and discriminative features is a crucial component of ReID, 
and has been dominated by CNN-based methods for a long time \cite{khorramshahi2020devil,PCB, Circle_loss, MGN, guangcongwang}. 

By reviewing CNN-based  methods, we find two important issues which are not well addressed in the field of object ReID. (1) Exploiting the \textit{rich structural patterns in a  global scope} is crucial for object ReID \cite{RGA}. However, CNN-based methods mainly focus on small discriminative regions due to a Gaussian distribution of effective receptive fields \cite{receptive_field}. 
Recently, attention modules~\cite{RGA, SCSN, ABD-Net,SONA, HACNN, MHN} have been introduced to explore long-range dependencies \cite{non_local}, but most of them are embedded in the deep layers and do not solve the principle problem of CNN. Thus, attention-based methods still prefer large continuous areas and are hard to extract multiple diversified discriminative parts (see Figure \ref{fig:motivation}). 
(2) Fine-grained features with \textit{detail information} are also important. However, the downsampling operators (\eg pooling and strided convolution) of CNN reduce spatial resolution of output feature maps, which greatly affect the discrimination ability to distinguish objects with similar appearances~\cite{PCB, luo2019bag}. As shown in Figure \ref{fig:feature_map}, the details of the backpack are lost in CNN-based feature maps, making it difficult to differentiate the two people.

Recently, Vision Transformer (ViT) \cite{ViT} and Data-efficient image Transformers (DeiT) \cite{touvron2020training} have shown that pure transformers can be as effective as CNN-based methods on feature extraction for image recognition. With the introduction of multi-head attention modules and the removal of convolution and downsampling operators, transformer-based models are suitable to solve the aforementioned problems in CNN-based ReID for the following reasons.  (1) The multi-head self-attention captures long range dependencies and drives the model to attend diverse human-body parts than CNN models (\eg thighs, shoulders, waist in Figure~\ref{fig:motivation}).
 (2) Without downsampling operators, transformer can keep more detailed information. For example, one can observe that the difference on feature maps around backpacks (marked by red boxes in Figure~\ref{fig:feature_map}) can help the model easily differentiate the two people. 
 These advantages motivate us to introduce pure transformers in the object ReID.

Despite its great advantages as discussed above, transformers still need to be designed specifically for object ReID to tackle the unique challenges, such as the large variations (\eg occlusions, diversity of poses, camera perspective) in images. 
Substantial efforts have been devoted to alleviating this challenge in CNN-based methods. Among them, local part features \cite{PCB, MGN,MSCAN,DPL, luo2019alignedreid++} and side information (such as cameras and viewpoints) \cite{chu2019vehicle, camera-bn,personX, PVEN}, have been proven to be essential and effective to enhance the feature robustness. Learning part/stripe aggregated features makes it robust against occlusions and misalignments~\cite{reid_survey}. However, extending the rigid stripe part methods from CNN-based methods to pure transformer-based methods may damage long-range dependencies due to global sequences splitting into several isolated subsequences. In addition, taking side information into consideration, such as camera and viewpoint-specific information, an invariant feature space can be constructed to diminish bias brought by side information variations. However, the complex designs for side information built on CNN, if directly applied to transformers, cannot make full use of the inherent encoding capabilities of transformers. As a result, specific designed modules are inevitable and essential for a pure transformer to successfully handle these challenges.

Therefore, we propose a new object ReID framework dubbed TransReID to learn robust feature representations. Firstly, by making several critical adaptations, we construct a strong baseline framework based on a pure transformer. 

Secondly, in order to expand long-range dependencies and enhance feature robustness, we propose a \textit{jigsaw patches module} (JPM) by rearranging the patch embeddings via shift and shuffle operations and re-grouping them for further feature learning. The JPM is employed on the last layer of the model to extract robust features in parallel with the global branch which does not include this special operation. Hence, the network tends to extract perturbation-invariant and robust features with global context. Thirdly, to further enhance the learning of robust features, a \textit{side information embedding} (SIE) is introduced. Instead of the special and complex designs in CNN-based methods for utilizing these non-visual clues,
we propose a unified framework that effectively incorporates non-visual clues through learnable embeddings to alleviate the data bias brought by cameras or viewpoints. Taking cameras for example, the proposed SIE helps address the vast pairwise similarity discrepancy between inter-camera and intra-camera matching (see Figure \ref{fig:bias}). 
SIE can also be easily extended to include any non-visual clues other than the ones we have demonstrated.

To our best knowledge, we are the first to investigate the application of pure transformers in the field of object ReID. The contributions of the paper are summarised:
\begin{itemize}[noitemsep,nolistsep]
    \item We propose a strong baseline that exploits the pure transformer for ReID tasks for the first time and achieve comparable performance with CNN-based frameworks.
    \item We design a \textit{jigsaw patches module} (JPM), consisting of shift and patch shuffle operation, which facilitates perturbation-invariant and robust feature representation of objects.
    \item We introduce a \textit{side information embeddings} (SIE) that encodes side information by learnable embeddings, and is shown to effectively mitigate the bias of learned features.
    \item The final framework TransReID achieves state-of-the-art performance on both person and vehicle ReID benchmarks including MSMT17\cite{MSMT17}, Market-1501\cite{Market1501}, DukeMTMC-reID\cite{DukeMTMC-reID}, Occluded-Duke\cite{miao2019pose}, VeRi-776\cite{VeRi776} and VehicleID\cite{VehicleID}.
\end{itemize}

\section{Related Work}

\subsection{Object ReID}

The studies of object ReID have  been mainly focused on person ReID and vehicle ReID, with most state-of-the-art methods based on the CNN structure. A popular pipeline for object ReID is to design suitable loss functions to train a CNN backbone (e.g. ResNet~\cite{he2016deep}), which is used to extract features of images. The cross-entropy loss (ID loss)~\cite{zheng2018discriminatively} and triplet loss~\cite{liu2017end} are most widely used in the deep ReID. Luo \etal~\cite{luo2019bag} proposed the BNNeck to better combine ID loss and triplet loss. Sun \etal~\cite{Circle_loss} proposed a unified perspective for ID loss and triplet loss.

\textbf{Fine-grained Features.}  Fine-grained features have been learned to aggregate information from different part/region. The fine-grained parts are either automatically generated by roughly horizontal stripes or by semantic parsing. Methods like PCB \cite{PCB}, MGN \cite{MGN}, AlignedReID++ \cite{luo2019alignedreid++}, SAN \cite{qian2020stripe}, \etc, divide an image into several stripes and extract local features for each stripe. Using parsing or keypoint estimation to align different parts or two objects has also been proven effective for both person and vehicle ReID \cite{liu2020beyond,PVEN,wei2017glad,miao2019pose}.

\textbf{Side Information.} For images captured in a cross-camera system, large variations exist  in terms of  pose, orientation, illumination, resolution, \etc caused by different camera setup and object viewpoints. Some works \cite{camera-bn,chu2019vehicle} use side information such as camera ID or viewpoint information to learn invariant features. For example, Camera-based Batch Normalization (CBN) \cite{camera-bn} forces the image data from different cameras to be projected onto the same subspace, so that the distribution gap between inter- and intra- camera pairs is largely diminished. Viewpoint/Orientation-invariant feature learning \cite{chu2019vehicle,zhu2020aware} is also important for both person and vehicle ReID.

\subsection{Pure Transformer in Vision}

The Transformer model is proposed in \cite{transformer} to handle sequential data in the field of natural language processing (NLP). Many studies also show its effectiveness for computer-vision tasks. Han \etal~\cite{han2020survey} and Salman \etal~\cite{khan2021transformers} have surveyed the application of the Transformer in the field of computer vision.

Pure Transformer models are becoming more and more popular. For example, Image Processing Transformer (IPT)~\cite{IPT} takes advantage of transformers by using large scale pre-training and achieves the state-of-the-art performance on several image processing tasks like super-resolution, denoising and de-raining. ViT \cite{ViT} is proposed recently  which applies a pure transformer directly to sequences of image patches. However, ViT requires a large-scale dataset to pretrain the model. To overcome this shortcoming, Touvron \etal \cite{touvron2020training} propose a framework called DeiT which introduces a teacher-student strategy specific for transformers to speed up  ViT training without the requirement of large-scale pretraining data.

\section{Methodology}
\begin{figure*}[ht]
    \vspace{-1.5em}
\begin{center}
   \includegraphics[width=0.95\linewidth]{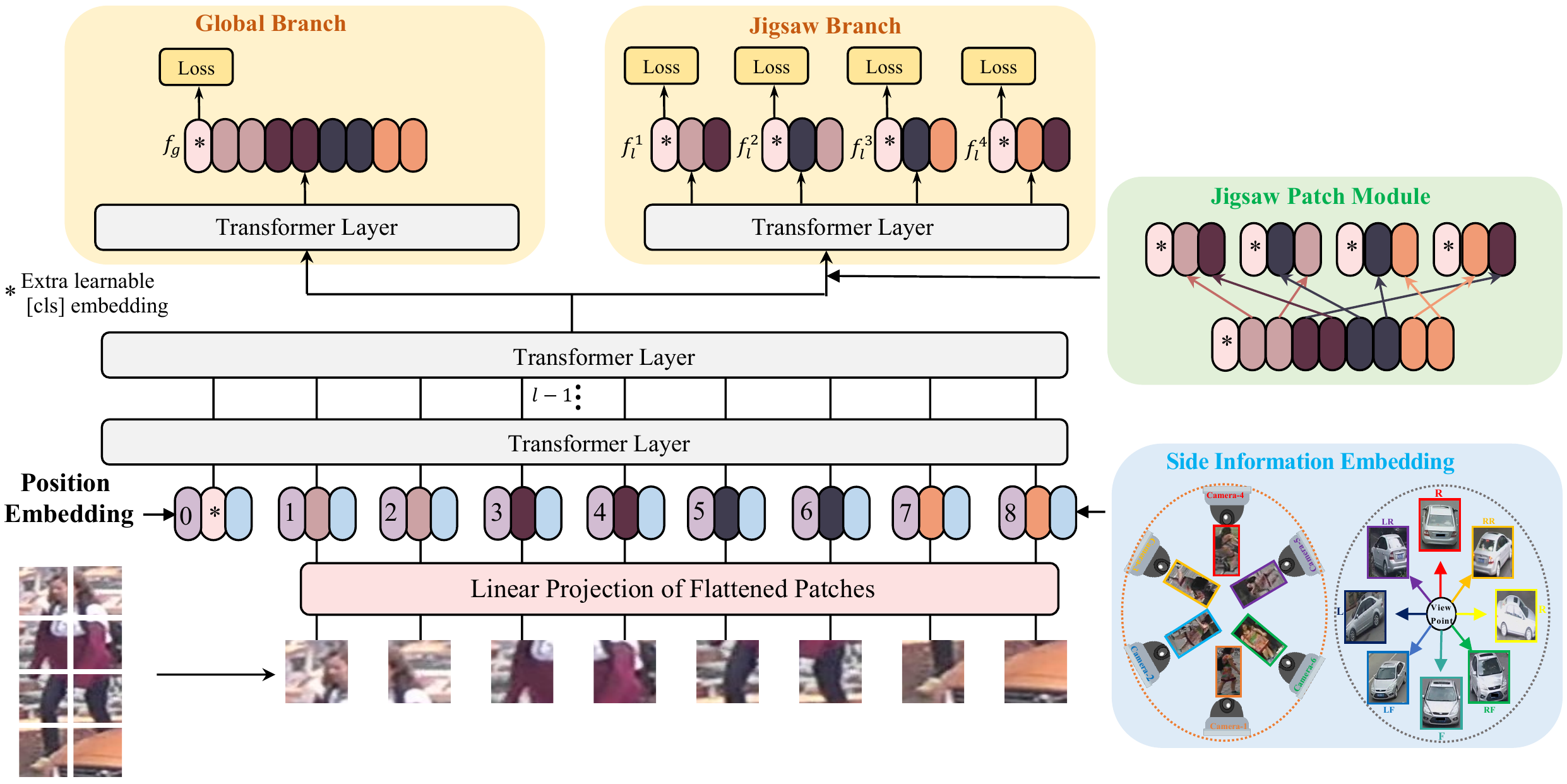}
\end{center}
    \vspace{-1em}
   \caption{Framework of proposed TransReID. Side Information Embedding (light blue) encodes non-visual information such as camera or viewpoint into embedding representations. It is input into transformer encoder together with patch embedding and position embedding. Last layer includes two independent transformer layers. One is standard to encode global feature. The other contains the Jigsaw Patch Module (JPM) which shuffles all patches and regroups them into several groups. All these groups are input into a shared transformer layer to learn local features. Both global feature and local features contribute to ReID loss.}
\label{fig:framework}
\end{figure*}

Our object ReID framework is based on transformer-based image classification, but with several critical improvements to capture robust feature (Sec.~\ref{ssec:strong_baseline}). To further boost the robust feature learning in the context of transformer, a jigsaw patch module (JPM) and a side information embeddings (SIE) are carefully devised in Sec.~\ref{ssec:jpm} and Sec.~\ref{ssec:sie}. The two modules are jointly trained in an end-to-end manner and shown in Figure~\ref{fig:framework}.

\subsection{Transformer-based strong baseline}
\label{ssec:strong_baseline}
\begin{figure}[ht]
\begin{center}
   \includegraphics[width=1\linewidth]{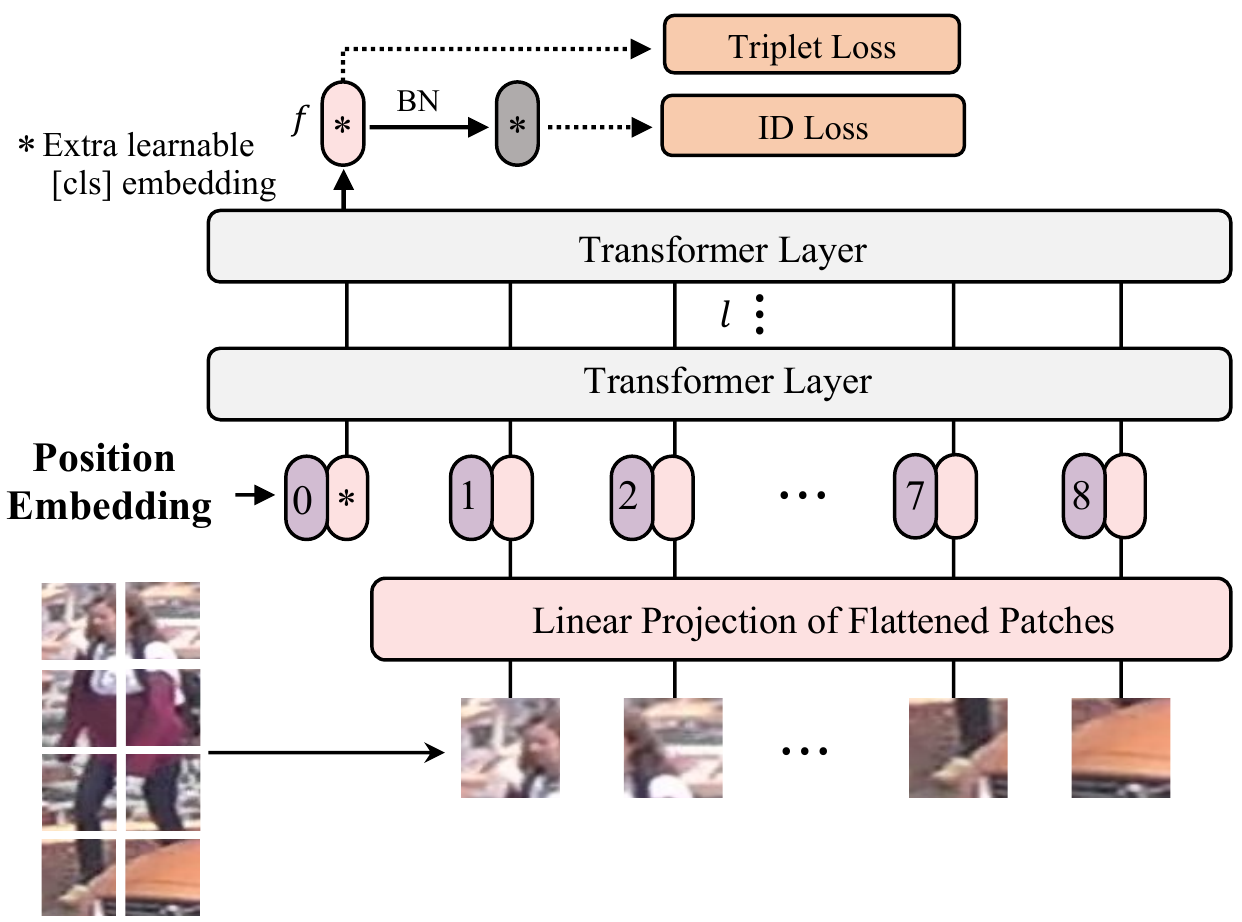}
\end{center}
    \vspace{-1.em}
   \caption{Transformer-based strong baseline framework (a non-overlapping partition is shown). Output [cls] token marked with $*$ is served as the global feature $f$. Inspired by \cite{luo2019bag}, we introduce the BNNeck after the $f$.}
  \label{fig:strong_baseline}
\end{figure}

We build a transformer-based strong baseline for object ReID, following the general strong pipeline for object ReID~\cite{luo2019bag, MGN}. Our method has two main stages, \ie, feature extraction and supervision learning. 
As shown in Figure~\ref{fig:strong_baseline}. Given an image $x \in \mathbb{R}^{H \times W \times C}$, where $H$, $W$, $C$ denote its height, width, and number of channels, respectively, we split it into $N$ fixed-sized patches $\{x^i_p|i=1,2,\cdots, N\}$. An extra learnable [cls] embedding token denoted as $x_\text{cls}$ is prepended to the input sequences. The output [cls] token serves as a global feature representation $f$. 
Spatial information is incorporated by adding learnable position embeddings. Then, the input sequences fed into transformer layers can be expressed as:
\begin{equation}
    \mathcal{Z}_0 = [ {x}_\text{cls}; \, \mathcal{F}({x}^1_p) ; \, \mathcal{F}({x}^2_p); \cdots; \, \mathcal{F}({x}^{N}_p) ] + \mathcal{P},
\end{equation}
where $\mathcal{Z}_0$ represents input sequence embeddings and $\mathcal{P} \in \mathbb{R}^{(N + 1) \times D}$ is position embeddings. $\mathcal{F}$ is a linear projection mapping the patches to $D$ dimensions.
Moreover, $l$ transformer layers are employed to learn feature representations. 
The limited receptive field problem of CNN-based methods is addressed, because all transformer layers have a global receptive field. There are also no downsampling operations, so the detailed information is preserved.

\textbf{Overlapping Patches.} 
Pure transformer-based models (\eg ViT, DeiT) split the images into non-overlapping patches, losing local neighboring structures around the patches. 
Instead, we use a sliding window to generate patches with overlapping pixels. Denoting the step size as $S$, size of the patch as $P$ (\eg $16$) , then the shape of the area where two adjacent patches overlap is $(P-S)\times P$. An input image with a resolution $H\times W$ will be split into $N$ patches.
\begin{equation}\label{eq:raw_input}
N=N_H \times N_W = \lfloor \frac{H+S-P}{S} \rfloor \times \lfloor \frac{W+S-P}{S} \rfloor  
\end{equation}
where $ \lfloor \cdot \rfloor$ is the floor function and $S$ is set smaller than $P$. $N_H$ and $N_W$ represent the numbers of splitting patches in height and width, respectively. 
The smaller $S$ is, the more patches the image will be split into. 
Intuitively, more patches usually bring better performance with the cost of more computations. 

\textbf{Position Embeddings.} As the image resolution for ReID tasks may be different from the original one in image classification, the position embedding pretrained on ImageNet cannot be directly loaded here. Therefore, a bilinear 2D interpolation is introduced to help handle any given input resolution. Similar to ViT, the position embedding is also learnable.

\textbf{Supervision Learning.} 
We optimize the network by constructing ID loss and triplet loss for global features. The ID loss $\mathcal{L}_{ID}$ is the cross-entropy loss without label smoothing. For a triplet set $\{a, p ,n \}$, the triplet loss $\mathcal{L}_{T}$ with soft-margin is shown as follows: 
\begin{equation}
\mathcal{L}_{T}=\log \left[1+\exp \left(\left\|f_{a}-f_{p}\right\|_{2}^{2}-\left\|f_{a}-f_{n}\right\|_{2}^{2} \right)\right]
\end{equation}


\subsection{Jigsaw Patch Module}\label{ssec:jpm}

Although transformer-based strong baseline can achieve impressive performance in object ReID, it utilizes information from the entire image for object ReID. However, due to challenges like occlusions and misalignments, we may only have partial observation of an object. Learning fine-grained local features such as striped features has been widely used for CNN-based methods to tackle these challenges. 

Suppose the hidden features input to the last layer are denoted as $\mathcal{Z}_{l-1} = [z^0_{l-1};z^1_{l-1},z^2_{l-1},...,z^N_{l-1}]$. To learn fine-grained local features, a straightforward solution is splitting $[z^1_{l-1},z^2_{l-1},...,z^N_{l-1}]$ into $k$ groups in order which concatenate the shared token $z^0_{l-1}$ and then feed $k$ feature groups into a shared transformer layer to learn $k$ local features denoted as $\{f^j_l|j=1,2,\cdots, k\}$ and $f^j_l$ is the output token of $j$-th group.
But it may not take full advantage of global dependencies for the transformer because each local segment only considers a part of the continuous patch embeddings. 

To address the aforementioned issues, we propose a jigsaw patch module (JPM) to shuffle the patch embeddings and then re-group them into different parts, each of which contains several random patch embeddings of an entire image. In addition, extra perturbation introduced in training also helps improve the robustness of object ReID model.
Inspired by ShuffleNet \cite{zhang2018shufflenet}, the patch embeddings are shuffled via a shift operation and a patch shuffle operation. The sequences embeddings $\mathcal{Z}_{l-1}$ are shuffled as follow:
\begin{itemize}
\item \textbf{Step1: The shift operation.} The first $m$ patches (except for [cls] token) are moved to the end, \ie $[z^1_{l-1},z^2_{l-1},...,z^N_{l-1}]$ is shifted in $m$ steps to become $[z^{m+1}_{l-1},z^{m+2}_{l-1},...,z^N_{l-1},z^1_{l-1},z^2_{l-1},...,z^m_{l-1}]$.
\item \textbf{Step2: The patch shuffle operation.} The shifted patches are further shuffled by the patch shuffle operation with $k$ groups. The hidden features become $[z^{x1}_{l-1},z^{x2}_{l-1},...,z^{x_N}_{l-1}], x_i \in [1,N]$.
\end{itemize}

With the shift and shuffle operation, the local feature $f^j_l$ can cover patches from different body or vehicle parts which means that the local features hold global discriminative capability.

As shown in Figure \ref{fig:framework}, paralleling with the jigsaw patch, another global branch which is a standard transformer encodes $\mathcal{Z}_{l-1}$ into $\mathcal{Z}_{l} = [f_g;z^1_{l},z^2_{l},...,z^N_{l}]$, where $f_g$ is served as the global feature of CNN-based methods. Finally, the global feature $f_g$ and $k$ local features are trained with $\mathcal{L}_{ID}$ and $\mathcal{L}_{T}$. The overall loss is computed as follow:

\vspace{-1.em}
\begin{equation}
\mathcal{L}=\mathcal{L}_{ID}(f_g)+\mathcal{L}_{T}(f_g)+ \frac{1}{k} \sum_{j=1}^k(\mathcal{L}_{ID}(f_l^j)+\mathcal{L}_{T}(f_l^j))
\end{equation}
\vspace{-1.em}

During inference, we concatenate the global feature and local features $[f_g,f_l^1,f_l^2,...,f_l^k]$ as the final feature representation. Using $f_g$  only is a variation with lower computational cost and slight performance degradation.


\subsection{Side Information Embeddings}\label{ssec:sie}

After obtaining fine-grained feature representations, features are still susceptible to camera or viewpoint variations. In other words, the trained model may easily fail to distinguish the same object from different perspectives due to scene-bias.
Therefore, we propose a Side Information Embedding (SIE) to incorporate the non-visual information, such as cameras or viewpoints, into embedding representations to learn invariant features.

Inspired by position embeddings which encode positional information adopting learnable embeddings, we plug learnable 1-D embeddings to retain side information. Particularly, as illustrated in Figure~\ref{fig:framework}, SIE is inserted into the transformer encoder together with patch embeddings and position embeddings.  
In specific, suppose there are $N_C$ camera IDs in total, we initialize learnable side information embeddings as $\mathcal{S}_C \in \mathbb{R}^{N_C \times D}$. If camera ID of an image is $r$, then its camera embeddings can be denoted as $\mathcal{S}_C[r]$. Different from the position embeddings which vary between patches, camera embeddings $\mathcal{S}_C[r]$ are the same for all patches of an image. In addition, if viewpoint of the object is available, either by a viewpoint estimation algorithm or human annotations, we can also encode the viewpoint label $q$ as $\mathcal{S}_V[q]$ for all patches of an image where $\mathcal{S}_V \in \mathbb{R}^{N_V \times D}$ and $N_V$ represents the number of viewpoint IDs.

Now comes the problem about how to integrate two different types of information. A trivial solution might be directly adding the two embeddings together like $\mathcal{S}_C[r] + \mathcal{S}_V[q]$. 
However, it might make the two embeddings counteract each other due to redundant or adversarial information. We propose to encode the camera and viewpoint jointly as  $\mathcal{S}_{(C,V)} \in \mathbb{R}^{(N_C \times N_V) \times D}$. 

Finally, the input sequences with camera ID $r$ and viewpoint ID $q$ are fed into transformer layers as follows:

\vspace{-0.5em}
\begin{equation}
    \mathcal{Z}_0^{'} = \mathcal{Z}_0 + \lambda \mathcal{S}_{(C,V)}[r*{N_k}+q],
\end{equation}
where $\mathcal{Z}_0$ is the raw input sequences in Eq.~\ref{eq:raw_input} and $\lambda$ is a hyperparameter to balance the weight of SIE. As the position embeddings are different for each patch but the same across different images, and $\mathcal{S}_{(C,V)}$ are the same for each patch but may have different values for different images. Transformer layers are able to encode embeddings with different distribution properties which can then be added directly.

Here we have only demonstrate the usage of SIE with camera and viewpoint information which are both categorical variables. In practice, SIE can be further extended to encode more kinds of information, including both categorical and numerical variables. In our experiments on different benchmarks, camera and viewpoint information is included wherever available.

\section{Experiments}

 \subsection{Datasets}
 We evaluate our proposed method on four person ReID datasets, Market-1501 \cite{Market1501}, DukeMTMC-reID \cite{DukeMTMC-reID}, MSMT17 \cite{MSMT17}, Occluded-Duke \cite{miao2019pose}, and two vehicle ReID datasets, VeRi-776 \cite{VeRi776} and VehicleID \cite{VehicleID}. It is noted that, unlike other datasets, images in Occluded-Duke are selected from DukeMTMC-reID and the training/query/gallery set contains 9\%/ 100\%/ 10\% occluded images respectively. All datasets except VehicleID provide camera ID for each image, while only VeRi-776 and VehicleID dataset provide viewpoint labels for each image. The details of these datasets are summarized in Table \ref{tab:dataset}.

\begin{table}[ht]
    \centering
    \footnotesize
        \begin{tabular}{ l|ccccc}
        \hline
        Dataset   & Object    & \#ID   & \#image    & \#cam  & \#view  \\
        \hline
        MSMT17  &Person & 4,101  & 126,441    &15 & -   \\
        Market-1501  &Person & 1,501  & 32,668    &6 & -   \\
        DukeMTMC-reID  &Person & 1,404  & 36,441    &8 & -   \\
        Occluded-Duke  &Person & 1,404  & 36,441    &8 & -   \\
        VeRi-776  &Vehicle & 776  & 49,357     &20 & 8   \\
        VehicleID  &Vehicle & 26,328  & 221,567    &- & 2   \\
        \hline
        \end{tabular}
    \caption{ Statistics of datasets used in the paper.}                    \label{tab:dataset}
\end{table}

 \subsection{Implementation}
Unless otherwise specified, all person images are resized to $256 \times 128$ and all vehicle images are resized to $256 \times 256 $. The training images are augmented with random horizontal flipping, padding, random cropping and random erasing \cite{random_erase3}. The batch size is set to 64 with 4 images per ID. SGD optimizer is employed with a momentum of 0.9 and the weight decay of 1e-4. The learning rate is initialized as 0.008 with cosine learning rate decay. Unless otherwise specified, we set $m=5, k=4$ and $m=8, k=4$ for person and vehicle ReID datasets, respectively.

All the experiments are performed with one Nvidia Tesla V100 GPU using the PyTorch toolbox \footnote{http://pytorch.org} with FP16 training . The initial weights of ViT are pre-trained on ImageNet-21K and then finetuned on ImageNet-1K, while the initial weights of DeiT are trained only on ImageNet-1K.

\textbf{Evaluation Protocols.} Following conventions in the ReID community, we evaluate all methods with Cumulative Matching Characteristic (CMC) curves and the mean Average Precision (mAP).

\subsection{Results of Transform-based Baseline}

\renewcommand{\multirowsetup}{\centering}
\begin{table}[t]
\footnotesize
    \begin{center}
    \begin{tabular}{ cc|cc|cc}
    \hline
    & Inference& \multicolumn{2}{c|}{MSMT17} & \multicolumn{2}{c}{VeRi-776} \\
    Backbone      & Time  & mAP    & R1  & mAP & R1  \\
    \hline
    \hline
    ResNet50    & 1x      &   51.3    & 75.3  &  76.4 & 95.2 \\
    ResNet101   & 1.48x   &   53.8    & 77.0  &  76.9 & 95.2 \\
    ResNet152   & 1.96x   &   55.6    & 78.4  &  77.1 & 95.9 \\
    ResNeSt50   & 1.86x   &   61.2    & 82.0  &  77.6 & 96.2 \\
    ResNeSt200  & 3.12x   &   63.5    & 83.5  &  77.9 & 96.4 \\
    \hline
    DeiT-S/16    & 0.97x   &   55.2    & 76.3  & 76.3  & 95.5 \\
    DeiT-B/16    & 1.79x   &    61.4  &  81.9 & 78.4  & 95.9  \\
    ViT-B/16     & 1.79x   &   61.0    & 81.8  &  78.2 & 96.5 \\
    ViT-B/16$_{s=14}$ & 2.14x   &   63.7    & 82.7  &  78.6 & 96.4\\
    ViT-B/16$_{s=12}$ & 2.81x   &   64.4    & 83.5  &  79.0 & 96.5 \\
    \hline
    \end{tabular}
    \end{center}
\caption{\label{tab:vit-bot} Comparison of different backbones. Inference time is represented by comparing each model to ResNet50 as only relative comparison is necessary. All the experiments were carried out on the same machine for fair comparison. \textbf{ViT-B/16 is regarded as the baseline model and abbreviated as Baseline in the rest of this paper}.}
\end{table}

In this section, we compare CNN-based and transformer-based backbones in Table \ref{tab:vit-bot}. To show the trade-off between computation and performance, several different backbones are chosen. DeiT-small, DeiT-Base, ViT-Base denoted as DeiT-S, DeiT-B, ViT-B, respectively. ViT-B/16$_{s=14}$ means ViT-Base with patch size 16 and step size $S=14$ in overlapping patches setting.
For a comprehensive comparison, inference time consumption of each backbone is included as well. 

We can observe a large gap in model capacity between the ResNet series and DeiT/ViT. DeiT-S/16 is a little bit better in performance and speed compared to ResNet50. DeiT-B/16 and ViT-B/16 achieve similar performance with ResNeSt50 \cite{zhang2020resnest} backbone, with less inference time than ResNeSt50 (1.79x vs 1.86x).  When we reduce the step size of the sliding window $S$, the performance of the Baseline can be improved while the inference time is also increasing. ViT-B/16$_{s=12}$ is faster than ResNeSt200 (2.81x vs 3.12x) and performs slightly better than ResNeSt200 on ReID benchmarks. Therefore, ViT-B/16$_{s=12}$ achieves better speed-accuracy trade-off than ResNeSt200. In addition, we believe that DeiT/ViT still have lots of room for improvement in terms of computational efficiency.

\renewcommand{\multirowsetup}{\centering}
\begin{table}[tb]
\footnotesize
    \begin{center}
    \begin{tabular}{ lc|cc|cc}
    \hline
    &  & \multicolumn{2}{c|}{MSMT17} & \multicolumn{2}{c}{VeRi-776} \\
    Backbone &\#groups    & mAP    & R1  & mAP & R1  \\
    \hline
    \hline
    Baseline   & -       & 61.0  & 81.8  & 78.2  & 96.5 \\
    +JPM        & 1     &62.9   & 82.5  & 78.6  & 97.0 \\
    +JPM        & 2     &62.8   & 82.1  & 79.1  & 96.4\\
    +JPM        & 4     &\textbf{63.6}   & \textbf{82.5}  & \textbf{79.2}  & \textbf{96.8} \\
    +JPM w/o rearrange & 4& 63.1  & 82.4  &  79.0  & 96.7 \\
    +JPM w/o local  & 4  & 63.5 & 82.5     & 79.1 & 96.6 \\
    \hline
    \end{tabular}
    \end{center}
    \vspace{-1.em}
    \caption{\label{tab:jpm} The ablation study of jigsaw patch module. `w/o rearrange' means the patch features are split into parts without rearrange including shift and shuffle operation. `w/o local' means we evaluate the global feature without concatenating local features.}
    \vspace{-1.5em}
\end{table}

\subsection{Ablation Study of JPM}

\begin{figure}[t]
    \centering
	\includegraphics[width=0.5\textwidth]{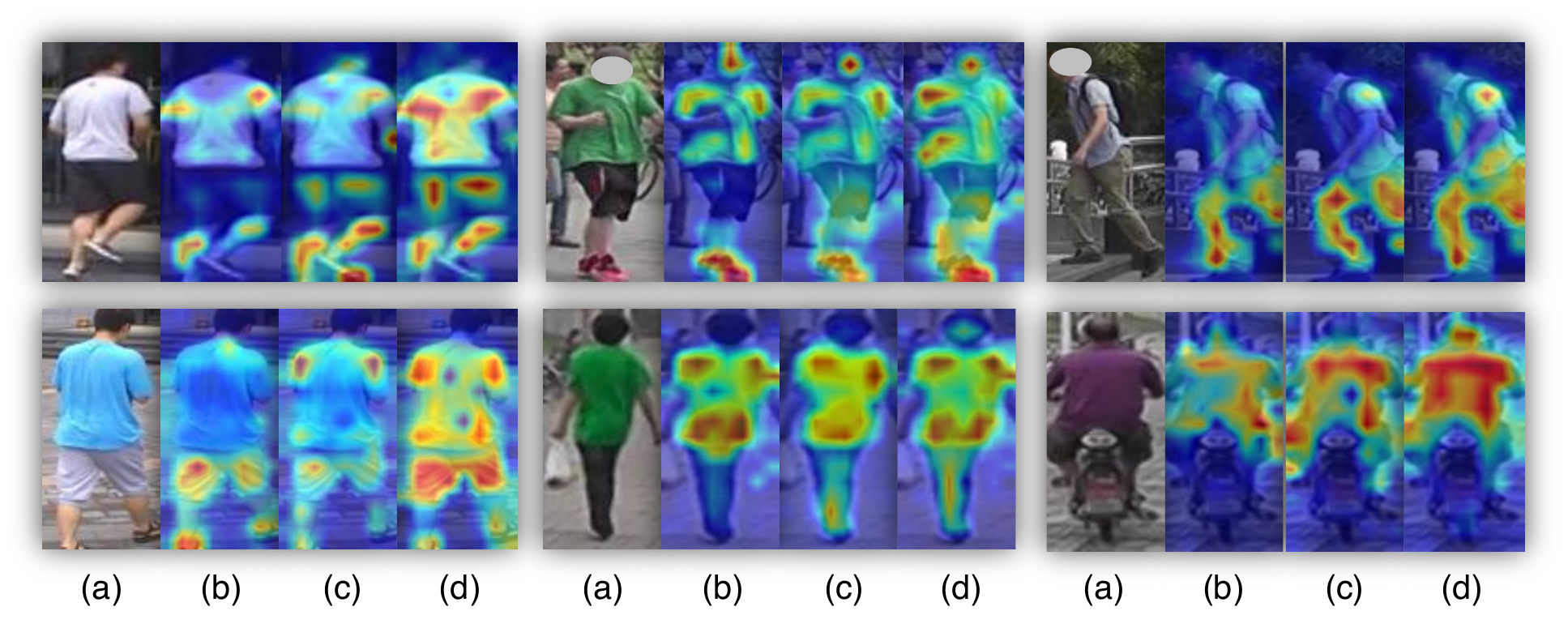}
	\vspace{-1.5em}
	\caption{Grad-CAM visualization of attention maps. (a) Input images, (b) Baseline, (c) JPM w/o rearrange, (d) JPM.} 
	\vspace{-1.em}
	\label{fig:jpm}
\end{figure}

The effectiveness of the proposed JPM module is validated in Table \ref{tab:jpm}. JPM provides +2.6\% mAP and +1.0\% mAP improvements compared to baseline on MSMT17 and VeRi-776, respectively. Increasing the number of groups $k$ can improve the performance while slightly increasing inference time. In our experiment, $k=4$ is a choice to trade off speed and performance. Comparing JPM and JPM w/o rearrange, we can observe that the shift and shuffle operation helps the model learn more discriminative features with +0.5\% mAP and +0.2\% mAP improvements on MSMT17 and VeRi-776, respectively. It is also observed that, if only the global feature $f_g$ is used in inference stage (still trained with full JPM), the performance (denoted as  ``w/o local'') is nearly comparable with the version of full set of features, which suggests us to only use the global feature as an efficient variation with  lower storage cost and computational cost in the inference stage. The attention maps visualized in Figure \ref{fig:jpm} show that JPM with the rearrange operation can help the model learn more global context information and more discriminative parts, which makes the model more robust to perturbations.

\renewcommand{\multirowsetup}{\centering}
\begin{table}[tb]
\footnotesize
    \begin{center}
    \begin{tabular}{ c|cc|cc|cc}
    \hline
    & \multicolumn{2}{c|}{} & \multicolumn{2}{c|}{MSMT17} & \multicolumn{2}{c}{VeRi-776} \\
    Method & Camera    & Viewpoint  & mAP    & R1  & mAP & R1  \\
    \hline
    \hline
    Baseline&  &        & 61.0  & 81.8  & 78.2  & 96.5\\
    + $S_C[r]$& $\surd$ & & \textbf{62.4}  & \textbf{81.9}  & 78.7&97.1\\
    + $S_V[q]$&  &$\surd$ & -     & -     &78.5&96.9\\
    + $S_{(C,V)}$ & $\surd$ &$\surd$  & - & - &\textbf{79.6} & \textbf{96.9}\\
    \hline
    \end{tabular}
    \end{center}
    \vspace{-0.5em}
    \caption{\label{tab:SIE} Ablation study of SIE. Since the person ReID datasets do not provide viewpoint annotations, viewpoint information can only be encoded in VeRi-776.}
    \vspace{-0.5em}
\end{table}

\begin{figure}[tp]
     \centering
     \begin{subfigure}[b]{0.49\linewidth}\centering
         \includegraphics[width=\linewidth]{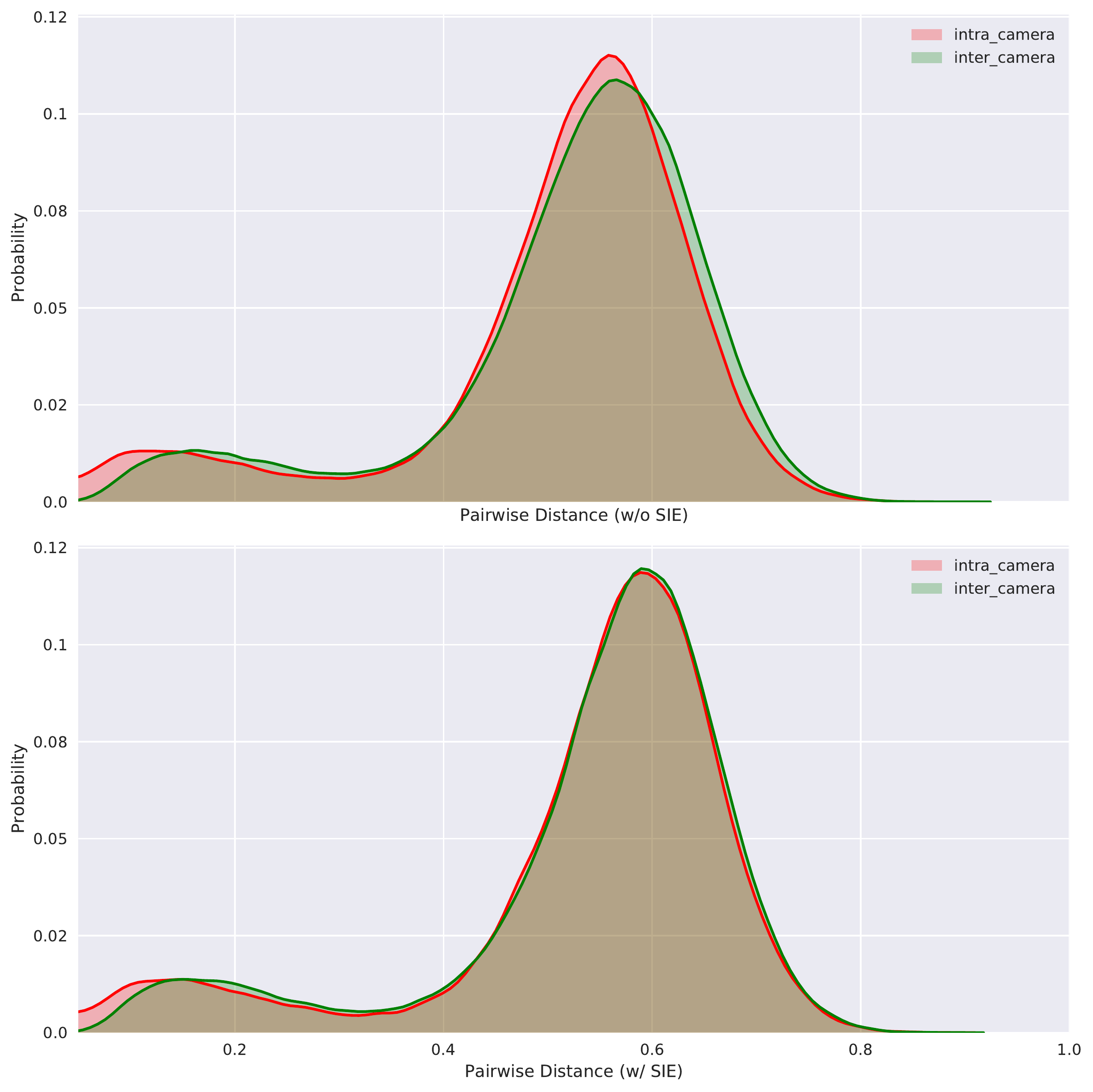}
         \caption{Distance of camera pairs.}
         \label{fig:a}
     \end{subfigure}
     \begin{subfigure}[b]{0.49\linewidth}\centering
         \includegraphics[width=\linewidth]{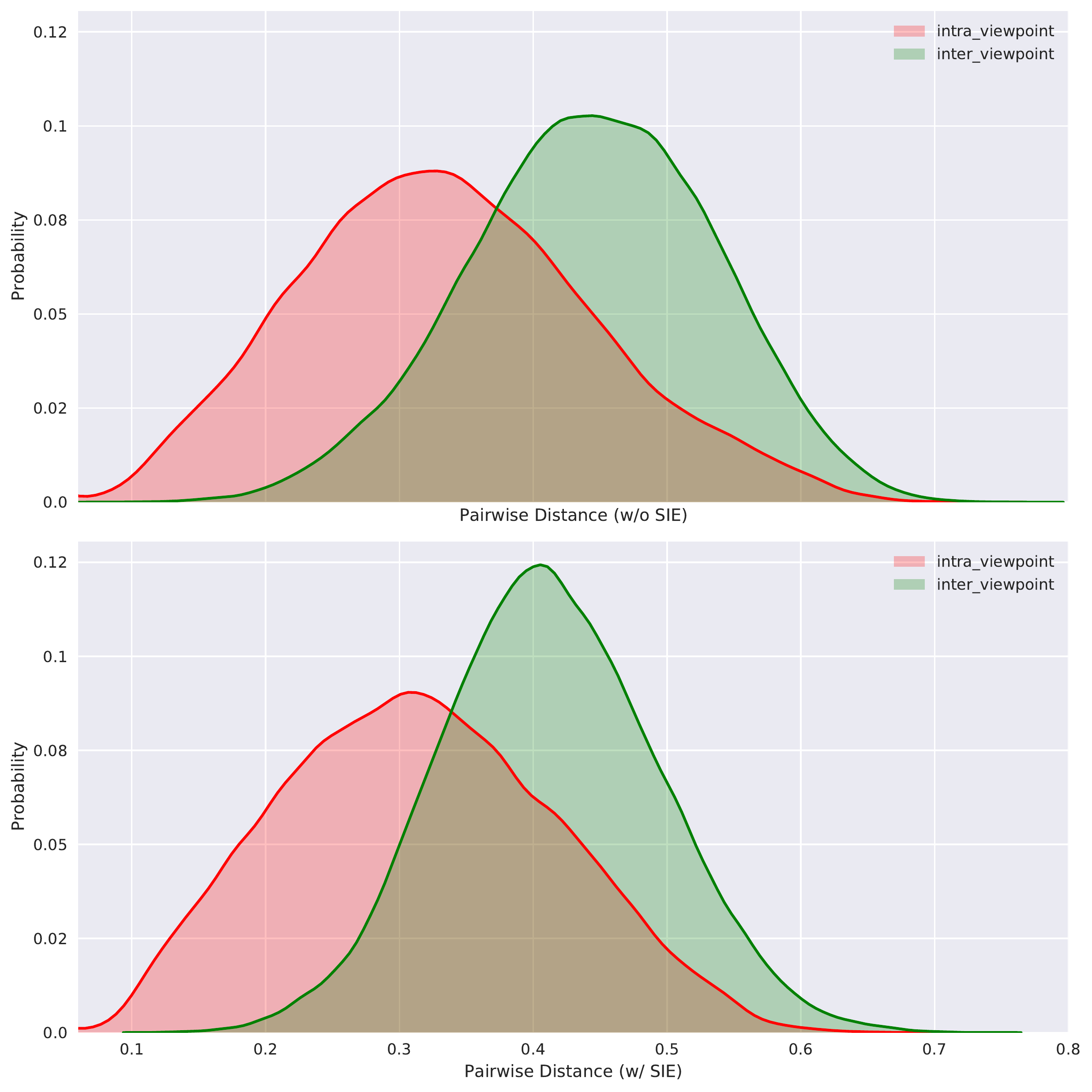}
         \caption{Distance of viewpoint pairs.}
         \label{fig:c}
     \end{subfigure}
     \\

    \caption{We visualize the distance distributions of different camera pairs and viewpoint pairs on VeRi-776. (a) inter-camera and intra-camera distance distribution. (b) inter-viewpoint and intra-viewpoint distance distribution.}
    \label{fig:bias}
\end{figure}

\begin{figure}[tp]
    \centering
    \begin{subfigure}[b]{0.49\linewidth}\centering
         \includegraphics[width=\linewidth]{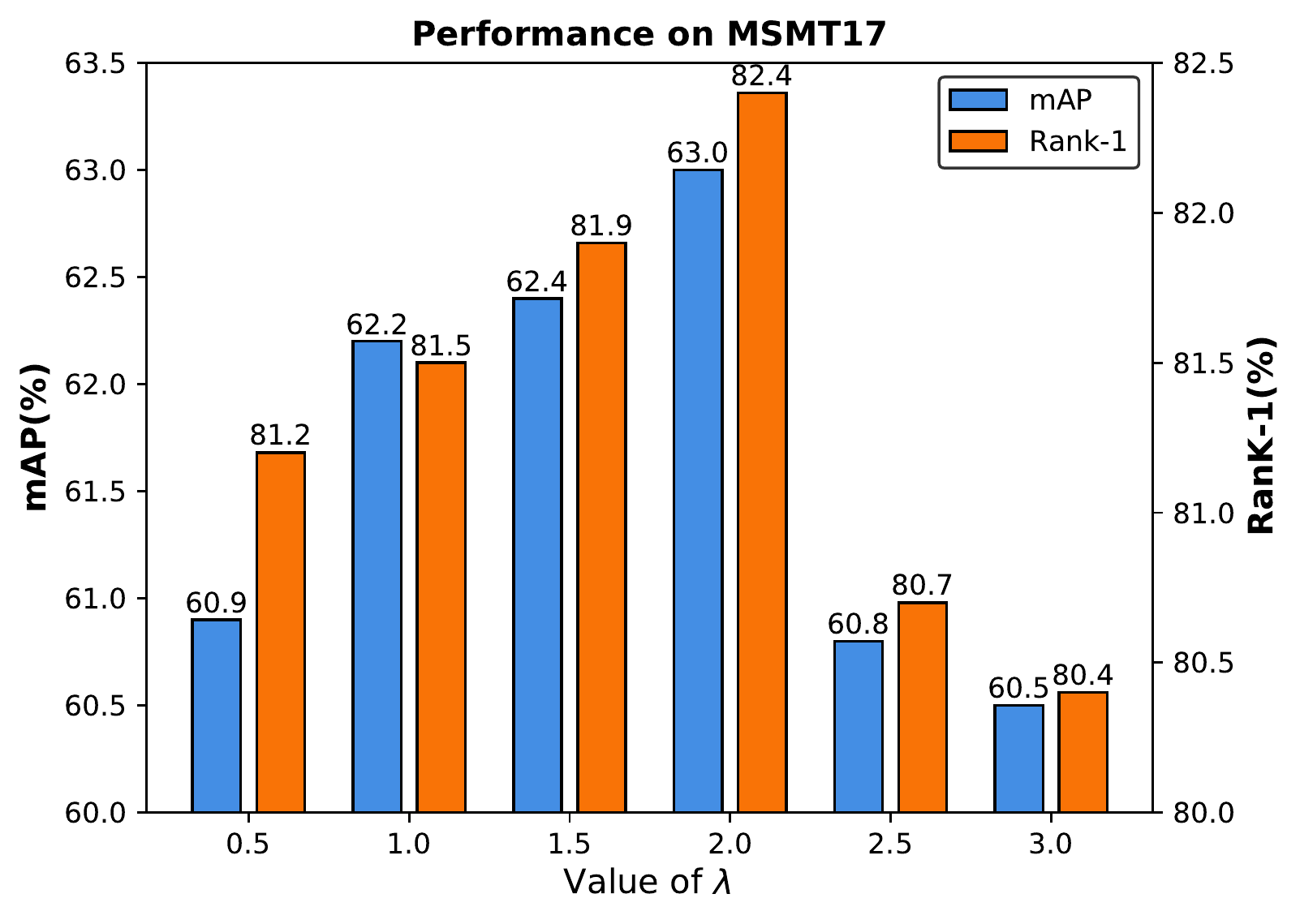}
         \caption{MSMT17}
         \label{fig:lambda1}
    \end{subfigure}
    \begin{subfigure}[b]{0.49\linewidth}\centering
         \includegraphics[width=\linewidth]{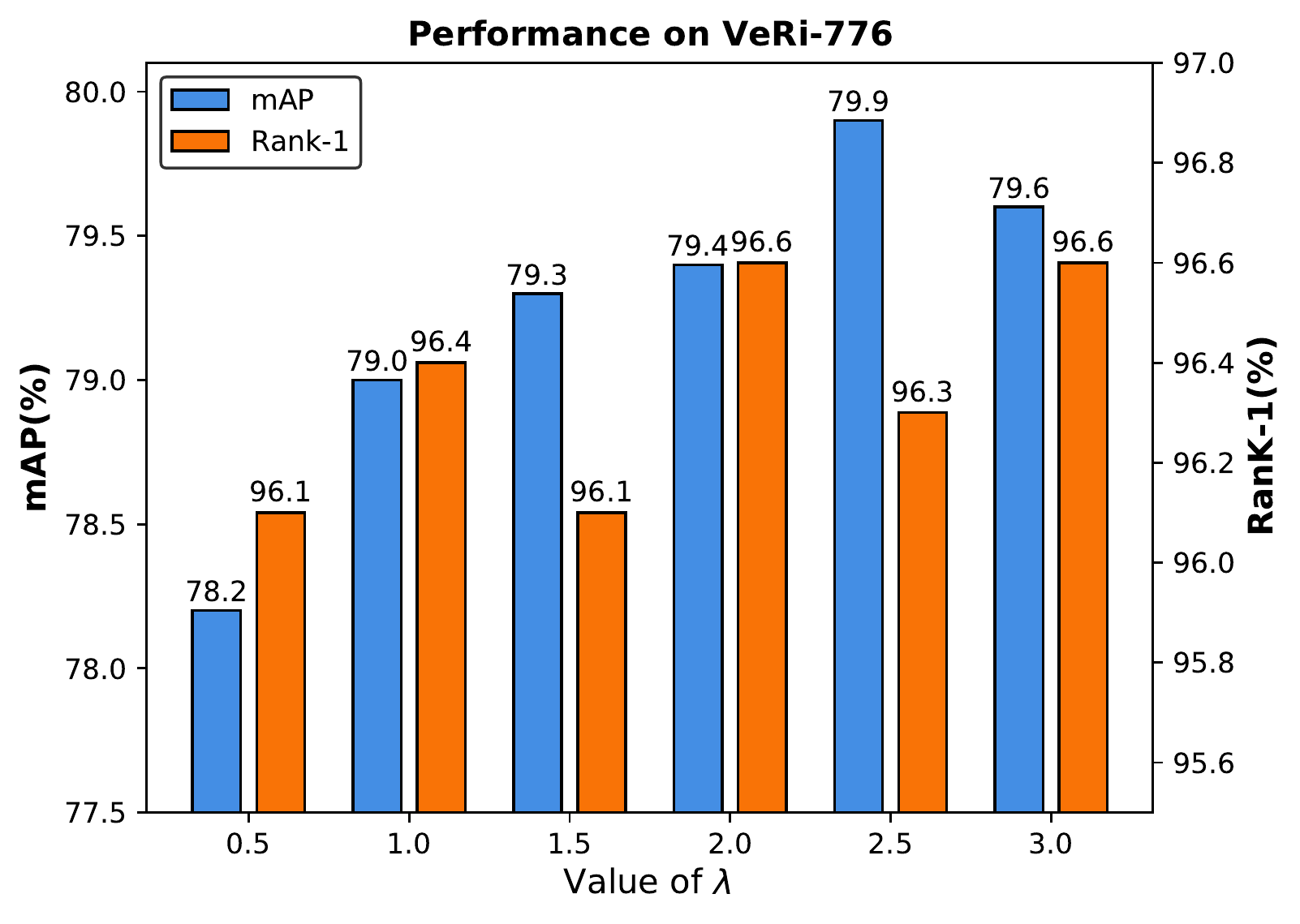}
         \caption{VeRi-776}
         \label{fig:lambda2}
    \end{subfigure}
    \caption{Impact of the hyper-parameter $\lambda$.}
    \vspace{-1em}
    \label{fig:lambda}
\end{figure}


\subsection{Ablation Study of SIE}

\textbf{Performance Analysis.} In Table~\ref{tab:SIE}, we evaluate the effectiveness of the SIE on MSMT17 and VeRi-776. MSMT17 does not provide viewpoint annotations, so the results of SIE which only encode camera information are shown for MSMT17. VeRi-776 not only have a camera ID of each image, but is also annotated with 8 different viewpoints according to vehicle orientation. Therefore, the results are shown with SIE encoding various combinations of camera ID and/or viewpoints information.

When SIE encodes only the camera IDs of images, the model gains 1.4\% mAP and 0.1\% rank-1 accuracy improvements on MSMT17. Similar conclusion can be made on VeRi-776. Baseline obtains 78.5\% mAP when SIE encodes viewpoint information. The accuracy increases to 79.6\% mAP when both camera IDs and viewpoint labels are encoded at the same time. If the encoding is changed to $\mathcal{S}_C[r] + \mathcal{S}_V[q]$, which is sub-optimal as discussed in Section \ref{ssec:sie}, we can only achieve 78.3\% mAP on VeRi-776. Therefore, the proposed $\mathcal{S}_{(C,V)}$ is a better encoding manner.

\textbf{Visualization of Distance Distribution.} 
As shown in Figure \ref{fig:bias}, the distribution gaps with cameras and viewpoints variations are obvious in Figure \ref{fig:a} and Figure \ref{fig:c}, respectively. When we introduce the SIE module into Baseline, the distribution gaps between inter-camera/viewpoint and intra-camera/viewpoint are reduced, which shows that the SIE module weakens the negative effect of the scene-bias caused by various cameras and viewpoints.

\textbf{Ablation Study of $\lambda$.} 
We analyze the influence of weight $\lambda$ of the SIE module on the performance in Figure \ref{fig:lambda}. When $\lambda = 0$, Baseline achieves 61.0\% mAP and 78.2\% mAP on MSMT17 and VeRi-776, respectively. With $\lambda$ increasing, the mAP is improved to 63.0\% mAP ($\lambda = 2.0$ for MSMT17) and 79.9\% mAP ($\lambda = 2.5$ for VeRi-776), which means the SIE module now is beneficial for learning invariant features. Continuing to increase $\lambda$, the performance is degraded because the weights for feature embedding and the position embedding are weakened.

\subsection{Ablation Study of TransReID}

Finally, we evaluate the benefits of introducing JPM and SIE in Table \ref{tab:transreid}. For the Baseline, JPM and SIE improve the performance by  +2.6\%/+1.0\% mAP and +1.4\%/+1.4\% mAP on MSMT17/VeRi-776, respectively. With these two modules used together, TransReID achieves 64.9\% (+3.9\%) mAP and 80.6\% (+2.4\%) mAP on MSMT17 and VeRi-776, respectively. 
The experimental results show the effectiveness of our proposed JPM, SIE, and the overall framework.


\renewcommand{\multirowsetup}{\centering}
\begin{table}[hb]
\footnotesize
    \begin{center}
    \begin{tabular}{ lcc|cc|cc}
    \hline
     & & &\multicolumn{2}{c|}{MSMT17} & \multicolumn{2}{c}{VeRi-776} \\
    Method   & JPM & SIE   & mAP    & R1  & mAP & R1  \\
    \hline
    \hline
    Baseline  & $\times$ & $\times$       & 61.0  & 81.8  & 78.2  & 96.5 \\
      & $\surd$ & $\times$             & 63.6  & 82.5  & 79.2  & 96.8 \\
    & $\times$     &$\surd$   & 62.4  & 81.9  &  79.6 & 96.9  \\
    TransReID  & $\surd$     &$\surd$      & \textbf{64.9}  &\textbf{83.3}  &\textbf{ 80.6}  &\textbf{96.9}  \\
    \hline
    \end{tabular}
    \vspace{-0.5em}
    \end{center}
    \caption{\label{tab:transreid} The ablation study of TransReID.}
    \vspace{-0.5em}
\end{table}

\renewcommand{\multirowsetup}{\centering}
\begin{table*}[ht]
\setlength\tabcolsep{4.5pt}
    \footnotesize
    \begin{center}
    \begin{tabular}{ccc|cccccccc||c|cccc}
    \hline
    \multicolumn{3}{c|}{} & \multicolumn{2}{c}{MSMT17} & \multicolumn{2}{c}{Market1501} & \multicolumn{2}{c}{DukeMTMC} & \multicolumn{2}{c||}{Occluded-Duke} & &\multicolumn{2}{c}{VeRi-776} &\multicolumn{2}{c}{VehicleID}\\
    Backbone & Method & Size    & mAP  & R1  & mAP & R1  & mAP  & R1  & mAP  & R1 & Method     &  mAP  & R1 & R1  & R5\\
    \hline
    \hline
    \multirow{7}{*}{CNN} &CBN$^{\tiny \textcircled{c}}$ \cite{camera-bn} 	& 256$\times$128    &42.9  & 72.8 & 77.3 & 91.3 & 67.3 & 82.5 & - &- & PRReID\cite{he2019part}  & 72.5&93.3 & 72.6& 88.6 \\
    &OSNet \cite{zhou2019omni} 		& 256$\times$128    &52.9  & 78.7 & 84.9 & 94.8 & 73.5 & 88.6 & - &- & SAN\cite{qian2020stripe}  & 72.5&93.3 & 79.7& 94.3 \\
    &MGN \cite{MGN}     & 384$\times$128    &52.1  & 76.9 & 86.9 & 95.7 & 78.4 & 88.7 & - &- & UMTS \cite{jin2020uncertainty}  & 75.9&95.8 & 80.9& 87.0 \\
    &RGA-SC \cite{RGA} & 256$\times$128    & 57.5 & 80.3 & 88.4 & \textbf{96.1} & - & - & - &- & VANet$^{\tiny \textcircled{v}}$ \cite{chu2019vehicle}  & 66.3& 89.8 & 83.3& 96.0 \\
    &SAN \cite{SAN}       & 256$\times$128  & 55.7 & 79.2 & 88.0 & 96.1 & 75.7 & 87.9 & - & - & SPAN $^{\tiny\textcircled{v}}$\cite{SPAN} & 68.9 & 94.0 & - & -\\
    &SCSN \cite{SCSN}       & 384$\times$128  & 58.5 & 83.8 & \textbf{88.5} & 95.7 & 79.0 & \textbf{91.0} & - &- & PGAN \cite{PGAN}& 79.3 & 96.5 & 78.0 & 93.2\\
    &ABDNet \cite{ABD-Net}       & 384$\times$128  &\textbf{ 60.8} & \textbf{82.3} & 88.3 & 95.6 & 78.6 & 89.0 & - &- & PVEN$^{\tiny\textcircled{v}}$ \cite{PVEN} & 79.5&95.6 & \textbf{84.7}& \textbf{97.0} \\
    &PGFA \cite{miao2019pose} & 256$\times$128  & - & - & 76.8 & 91.2 & 65.5 & 82.6  & 37.3 & 51.4 & SAVER \cite{khorramshahi2020devil} & 79.6 & 96.4 & 79.9& 95.2 \\
    &HOReID \cite{wang2020high} & 256$\times$128  & - & - & 84.9 & 94.2 & 75.6 & 86.9 & 43.8 & 55.1 & CFVMNet \cite{sun2020cfvmnet}  & 77.1&95.3 & 81.4& 94.1 \\
    &ISP \cite{ISP}       & 256$\times$128  & - & - & 88.6 & 95.3 & \textbf{80.0} & 89.6 & \textbf{52.3} &\textbf{62.8} & GLAMOR\cite{GLAMOR} & \textbf{80.3} & \textbf{96.5} & 78.6 & 93.6\\
    \hline
    \multirow{5}{*}{DeiT-B/16}     &Baseline & 256$\times$128 & 61.4 & 81.9 & 86.6  &94.4  &78.9 &89.3 &53.1 &60.6 
    &Baseline &78.4 &95.9 &83.1 &96.8\\
    
    &TransReID$^{\tiny\textcircled{c}}$ & 256$\times$128 & 63.9 & 82.7 & 88.0  &94.7  &81.2 &90.1 &55.6 &62.8 
    &TransReID $^{\tiny\textcircled{v}}$ &80.6 &96.8 &84.6 &97.4\\
    
    &TransReID$^{\tiny\textcircled{c}}$ & 384$\times$128 &65.5 &83.5&88.1 &94.9&81.3 &90.2&- &-  
    &TransReID$^{\tiny\textcircled{b}}$  & 81.2 & 96.8 & - & -\\
    
    &TransReID$^{*\tiny\textcircled{c}}$ & 256$\times$128 &\textbf{66.2} & \textbf{84.3} & \textbf{88.4} & \textbf{95.0} & \textbf{81.9} & \textbf{91.1} & \textbf{58.1} & \textbf{66.4}
    &TransReID$^{*\tiny \textcircled{v}}$&81.4 &96.8 &\textbf{85.2} &\textbf{97.6}  \\
    
    &TransReID$^{*\tiny \textcircled{c}}$ & 384$\times$128 &\textbf{66.3} &\textbf{84.5} &\textbf{88.5} &\textbf{95.1} &\textbf{82.1} &\textbf{91.1}&- &- &TransReID$^{*\tiny\textcircled{b}}$ &\textbf{82.3} &\textbf{97.1} &- &-  \\
    \hline
    
    \multirow{5}{*}{ViT-B/16} &Baseline & 256$\times$128 & 61.0 & 81.8 & 86.8  &94.7  &79.3 &88.8 &53.1 &60.5 
    &Baseline &78.2 &96.5 &82.3 &96.1\\
    &TransReID$^{\tiny\textcircled{c}}$ & 256$\times$128 & 64.9  &83.3  &88.2  &95.0  &80.6 &89.6 &55.7 &64.2 
    &TransReID $^{\tiny\textcircled{v}}$ &79.6 &97.0 &83.6&97.1\\
    &TransReID$^{\tiny\textcircled{c}}$ & 384$\times$128 & 66.6 & 84.6 & 88.8 & 95.0 & 81.8&90.4 &- &-  
    &TransReID$^{\tiny\textcircled{b}}$  &  80.6 & 96.9 &-&-\\
    &TransReID$^{*\tiny\textcircled{c}}$ & 256$\times$128 & \textbf{67.4} & \textbf{85.3} & \textbf{88.9} & \textbf{95.2} &\textbf{82.0} &\textbf{90.7} &\textbf{59.2} &\textbf{66.4}
    &TransReID$^{*\tiny \textcircled{v}}$  &  80.5 & 96.8 &\textbf{85.2}  & \textbf{97.5}\\
    &TransReID$^{*\tiny \textcircled{c}}$ & 384$\times$128 & \textbf{69.4} & \textbf{86.2} & \textbf{89.5} & \textbf{95.2} &\textbf{82.6} &\textbf{90.7} &- &- 
    &TransReID$^{*\tiny\textcircled{b}}$  &\textbf{82.0}  & \textbf{97.1}  & - & -\\
    \hline
    \end{tabular}
    \vspace{-1em}
    \end{center}
    \caption{\label{tab:sota} Comparison with state-of-the-art methods. DukeMTMC denotes the DukeMTMC-reID benchmark. The star * in the superscript means the backbone is with a sliding-window setting. Results are shown for person ReID datasets (left) and vehicle ReID datasets (right). Only the small subset of VehicleID is used in this paper. $^{\tiny \textcircled{c}}$ and $^{\tiny \textcircled{v}}$ indicate the methods are using camera IDs and viewpoint labels, respectively. $^{\tiny \textcircled{b}}$ means both are used. Viewpoint and camera information are used wherever available. Best results for previous methods and best of our methods are labeled in bold.}
\end{table*}

\subsection{Comparison with State-of-the-Art Methods}
In Table \ref{tab:sota}, our TransReID is compared with state-of-the-art methods on six benchmarks including person ReID, occluded ReID and vehicle ReID. 

\textbf{Person ReID.} 
On MSMT17 and DukeMTMC-reID, TransReID$^*$ (DeiT-B/16) outperforms the previous state-of-the-art methods by a large margin (+5.5\%/+2.1\% mAP). On Market-1501, TransReID$^*$ (256$\times$128) achieves comparable performance with state-of-the-art methods especially on mAP. Our method also shows superiority when compared with methods which also integrate camera information like CBN \cite{camera-bn}.

\textbf{Occluded ReID.} ISP implicitly uses human body semantic information through iterative clustering and HOReID introduces external pose models to align body parts. TransReID (DeiT-B/16) achieves 55.6\% mAP with a large margin improvement (at least +3.3\% mAP) compared to aforementioned methods, without requiring any semantic and pose information to align body parts, which shows the ability of TransReID to generate robust feature representations. Furthermore, TransReID$^*$ improves the performance to 58.1\% mAP with the help of overlapping patches.

\textbf{Vehicle ReID.} On VeRi-776, TransReID$^*$ (DeiT-B/16) reaches 82.3\% mAP surpassing GLAMOR by 2.0\% mAP. When only using viewpoint annotations, TransReID$^*$ still outperforms VANet and SAVER on both VeRi-776 and VehicleID. Our method achieves state-of-the-art performance about 85.2\% Rank-1 accuracy on VehicleID.

\textbf{DeiT vs ViT vs CNN.} TransReID$^*$ (DeiT-B/16) reaches competitive performance with existing methods under a fair comparison (ImageNet-1K pre-training). Extra results of our methods with ViT-B/16 are also reported in Table \ref{tab:sota} for further comparison. DeiT-B/16 achieves similar performance with ViT-B/16 for shorter image patch sequences. When the number of input patches is increasing, ViT-B/16 reaches better performance than DeiT-B/16, which shows ImageNet-21K pre-training provides ViT-B/16 better generalization capability.  Although CNN-based methods mainly report performance with the ResNet50 backbone, they may include multiple branches, attention modules, semantic models, or other modules that increase computational consumption. 
We have conducted a fair comparison on inference speed between TransReID$^*$ and MGN \cite{MGN} on the same computing hardware. Compared with MGN, TransReID* is 4.8\% faster in speed.
Therefore, TransReID* can achieve more promising performance under comparable computation to most of CNN-based methods.

\section{Conclusion}

In this paper, we investigate a pure transformer framework for the object ReID task, and propose two novel modules, \textit{i.e.,} jigsaw patch module (JPM) and side information embedding (SIE). The final framework TransReID outperforms all other state-of-the-art methods by a large margin on several popular person/vehicle ReID datasets including MSMT17, Market-1501, DukeMTMC-reID, Occluded-Duke, VeRi-776 and VehicleID.
Based on the promising results achieved by TransReID, we believe the transformer has great potential to be further explored for ReID tasks. Based on the rich experience gained from CNN-based methods, it is in prospect that more efficient transformer-based networks can be designed with better representation power and less computational cost.

\clearpage

{\small
\bibliographystyle{ieee_fullname}
\bibliography{egbib}
}

\clearpage

\appendix
\section*{Appendix}
\begin{table*}[b]
    \begin{center}\small
    \begin{tabular}{ c|c c c c c c c |cc|cc}
    \hline
  \multirow{2}{*}{Method} & \multirow{2}{*}{OPT} & \multirow{2}{*}{PE}  & \multirow{2}{*}{SP} & \multirow{2}{*}{DO} & \multirow{2}{*}{ADO} & \multirow{2}{*}{STL} & \multirow{2}{*}{LS}  &  \multicolumn{2}{c|}{MSMT17} & \multicolumn{2}{c}{VeRi-776} \\
    & & & & & & & & mAP  & R1  & mAP & R1  \\
    \hline
    \hline
     ViT-B/16 Baseline& SGD &  \cmark & \cmark & \xmark &  \xmark & \cmark & \xmark  & \textbf{61.0} & \textbf{81.8} & \textbf{78.2} & \textbf{96.5}\\ 
    \hline
    \multirow{2}{*}{Optimizer}& Adam &  \cmarkg & \cmarkg & \xmarkg & \xmarkg & \cmarkg & \xmarkg &  37.4 \blue{(-24.6)} &60.2 \blue{(-21.6)} & 65.8 \blue{(-12.4)} &91.7 \blue{(-4.8)} \\
    & AdamW &  \cmarkg & \cmarkg & \xmarkg & \xmarkg & \cmarkg & \xmarkg &  60.6 \blue{(-0.4)} & 81.7 \blue{(-0.1)} &78.0 \blue{(-0.2)} & 96.5 \blue{(-0.0)}\\
    \hline
    \multirow{4}{*}{{\makebox{\begin{minipage}{2cm}\centering Network\\Configuration \\ ~\end{minipage}}}}& \sgdg &  \xmark & \cmarkg & \xmarkg & \xmarkg & \cmarkg & \xmarkg &  22.4 \blue{(-38.6)} &38.3 \blue{(-43.5)} & 68.0 \blue{(-10.2)} & 92.8 \blue{(-3.7)}\\
    
    & \sgdg &  \cmarkg & \xmark & \xmarkg & \xmarkg & \cmarkg & \xmarkg & 59.9 \blue{(-1.1)} &80.2 \blue{(-1.6)} &  77.2 \blue{(-1.0)}& 96.1 \blue{(-0.4)} \\
    
    & \sgdg &  \cmarkg & \cmarkg & \cmark & \xmarkg & \cmarkg & \xmarkg  & 60.0 \blue{(-1.0)} & 80.7 \blue{(-1.1)} &78.0 \blue{(-0.2)}& 96.3 \blue{(-0.2)}\\
    & \sgdg &  \cmarkg & \cmarkg & \xmarkg & \cmark & \cmarkg & \xmarkg & 58.0 \blue{(-3.0)} & 78.8 \blue{(-3.0)} & 74.3 \blue{(-3.9)} & 94.9 \blue{(-1.6)}\\
    \hline
    \multirow{2}{*}{Loss Function}& \sgdg &  \cmarkg & \cmarkg & \xmarkg & \xmarkg & \xmark &\xmarkg & 60.3 \blue{(-0.7)} & 81.3 \blue{(-0.5)} & 77.5 \blue{(-0.7)} & 95.6 \blue{(-0.9)}\\
    & \sgdg &  \cmarkg & \cmarkg & \xmarkg & \xmarkg & \cmarkg & \cmark &  59.8 \blue{(-1.2)} & 80.4 \blue{(-1.4)} & 77.4 \blue{(-0.8)} & 96.5 \blue{(-0.0)}\\
   
    \hline
    \end{tabular}
    \end{center}
    \vspace{-0.5em}
    \caption{\label{tab:sb_analysis} Ablation study about training settings on MSMT17 and VeRi-776. The first row corresponds to the default configuration employed by our transformer-based strong baseline (ViT-B/16 as default backbones). The symbols \cmark and \xmark indicate that the corresponding setting is included or excluded, respectively. mAP(\%) and R1(\%) accuracy scores are reported. The abbreviations OPT, PE, SP, DO, ADO, STL, LS denote Optimizer, Position Embedding, Stochastic Depth \cite{stoc_depth}, Drop Out, Attention Drop Out, Soft Triplet Loss, Label Smoothing, respectively.}
    \vspace{-0.5em}
\end{table*}

\section{More Experimental Results}

\subsection{Study on Transformer-based Strong Baseline}
A transformer-based strong baseline with a few critical improvements has been introduced in Section~3.1 of the main paper. 
In this section, hyper-parameters and the settings for training such a baseline model will be analyzed in detail. Ablation studies are shown in Table \ref{tab:sb_analysis} for performance on MSMT17 and Veri-776 with different variations of the training settings.

\textbf{Initialization and hyper-parameters}.
 For our experiments, we initialize the pure transformer with ViT or DeiT ImageNet pre-trained weights and we initialize the weights for the SIE with a truncated normal distribution \cite{truncated_normal_dis}. Compared with ViT, DeiT is more sensitive to hyper-parameter settings. For the training of DeiT, we use a learning rate of 0.05 on MSMT17 and a high random erasing probability with 0.8 on each dataset to avoid overfitting. Other hyper-parameters settings are the same with ViT.

\textbf{Optimizer}.
Transformers are sensitive to the choice of the optimizer. Directly applying Adam optimizer with the hyper-parameters commonly used in ReID community \cite{luo2019bag} to transformer-based models will cause a significant drop in performance. AdamW~\cite{AdamW} is a commonly used optimizer for training transformer-based models, with much better performance compared with Adam. The best results are actually achieved by SGD in our experiments.

\textbf{Network Configuration}.
Position embeddings incorporate crucial spatial information which provides a significant boost in performance and is one of the key ingredients of our proposed training procedure. Without the position embeddings, the performance decreases by 38.6\% mAP and 10.2\% mAP on MSMT17 and VeRi-776, respectively.

Introducing  stochastic depth~\cite{stoc_depth} can boost the mAP performance by about 1\%, and it has also been proved to facilitate the convergence of transformer, especially for those deep ones~\cite{stoc_depth_1,stoc_depth_2}. 
Regarding other regularization methods, adding either drop out or attention drop out will result in performance drop. In our experiments, we set all the probability of regularization methods as 0.1.

\textbf{Loss Function}.
Different choices of loss functions have been compared in the bottom section of Table~\ref{tab:sb_analysis}. The soft version of triplet loss provides 0.7\% mAP improvement on MSMT17 compared with the regular triplet loss. Introducing label smoothing is harmful to performance, even though it has been a widely adopted trick. Therefore, the best combination for loss functions is soft triplet loss and cross entropy loss without label smoothing. 

\renewcommand{\multirowsetup}{\centering}
\begin{table}[t]
\footnotesize
    \begin{center}
    \begin{tabular}{ lc|cc|cc}
    \hline
    &  & \multicolumn{2}{c|}{MSMT17} & \multicolumn{2}{c}{VeRi-776} \\
    Backbone &\#groups    & mAP    & R1  & mAP & R1  \\
    \hline
    \hline
    Baseline (ViT-B/16)  & -       & 61.0  & 81.8  & 78.2  & 96.5 \\
    +JPM        & 1     &62.9   & 82.5  & 78.6  & 97.0 \\
    +JPM        & 2     &62.8   & 82.1  & 79.1  & 96.4\\
    +JPM        & 4     &\textbf{63.6}   & \textbf{82.5}  & \textbf{79.2}  & \textbf{96.8} \\
    +JPM w/o rearrange & 4& 63.1  & 82.4  &  79.0  & 96.7 \\
    +JPM w/o local  & 4  & 63.5 & 82.5     & 79.1 & 96.6 \\
    \hline
    Baseline (ViT-B/16$_{s=12}$) & -& 64.4  & 83.5  &  79.0 & 96.5 \\
    +JPM        & 4     & \textbf{66.5} &  \textbf{84.8}&   \textbf{80.0} & \textbf{97.0}   \\
    +JPM w/o rearrange & 4&  66.1 &  84.5 & 79.6  &  96.7\\  
    +JPM w/o local  & 4     &    66.3   &  84.5   &   79.8    & 96.8 \\
    \hline
    \end{tabular}
    \end{center}
    \vspace{-1.em}
    \caption{\label{tab:jpm-all} Detailed ablation study of jigsaw patch module (JPM). `w/o rearrange' means the patch sequences are split into subsequences without rearrangement. `w/o local' means we evaluate the global feature without concatenating local features.}
    \vspace{-1.5em}
\end{table}

\renewcommand{\multirowsetup}{\centering}
\begin{table}[t]
\footnotesize
    \begin{center}
    \begin{tabular}{c|cc|cc|cc}
    \hline
    & \multicolumn{2}{c|}{} & \multicolumn{2}{c|}{MSMT17} & \multicolumn{2}{c}{VeRi-776} \\
    Method & Camera & View & mAP    & R1  & mAP & R1  \\
    \hline
    \hline
    \multirow{4}{*}{\makebox{\begin{minipage}{2cm}\centering Baseline\\(ViT-B/16) \\ ~\end{minipage}}} & \xmarkg & \xmarkg       & 61.0  & 81.8  & 78.2  & 96.5\\
    & \cmark & \xmarkg & \textbf{62.4}  & \textbf{81.9}  & 78.7&97.1\\
    &  \xmarkg&\cmark & -     & -     &78.5&96.9\\
     & \cmark &\cmark  & - & - &\textbf{79.6} & \textbf{96.9}\\
    \hline
    \multirow{4}{*}{\makebox{\begin{minipage}{2cm}\centering Baseline\\(ViT-B/16$_{s=12}$) \\ ~\end{minipage}}}& \xmarkg & \xmarkg  &   64.4    & 83.5  &  79.0 & 96.5\\
                            & \cmark & \xmarkg    & \textbf{65.9}& \textbf{84.1}&79.4 &96.4\\   
                            &  \xmarkg & \cmark     & - & -&79.3 &97.0\\
                            & \cmark & \cmark  & - & - & \textbf{80.3}& \textbf{96.9}\\
    \hline
    \end{tabular}
    \end{center}
    \vspace{-0.5em}
    \caption{\label{tab:SIE-all} Detailed ablation study of side information embeddings (SIE). Experiments of viewpoint information are only conducted on VeRi-776 as the person ReID datasets do not provide viewpoint annotations. The symbols \cmark and \xmark indicate that the corresponding information is included or excluded.}
\end{table}

\subsection{More Ablation Studies of JPM and SIE}

In the main paper, we have demonstrated the effectiveness of using JPM and SIE based on the Baseline (ViT-B/16). More results about JPM and SIE are shown in Table~\ref{tab:jpm-all} and Table~\ref{tab:SIE-all} respectively, with the Baseline ViT-B/16$_{s=12}$,
which is supposed to have better feature representation ability and higher performance than ViT-B/16. From Table~\ref{tab:jpm-all}, we observe that: (1) The proposed JPM performs better with the rearrange schemes, indicating that the shift and patch shuffle operation help the model learn more discriminative features which are robust against perturbations. (2) The JPM module provides a consistent performance improvement over the baselines, no matter the baseline is ViT-B/16  or the stronger ViT-B/16$_{s=12}$,  demonstrating the effectiveness of the proposed JPM.

Similar conclusions can be made from Table~\ref{tab:SIE-all}. (1) We make better use of the viewpoint and camera information so that they are complementary with each other and combining them leads to the best performance. (2) Introducing SIE provides consistent improvement over the baselines of either ViT-B/16 or ViT-B/16$_{s=12}$.

\section {Analysis on Rearranging Patches in JPM}
\begin{figure}[t]
    \centering
    \includegraphics[width=0.5\textwidth]{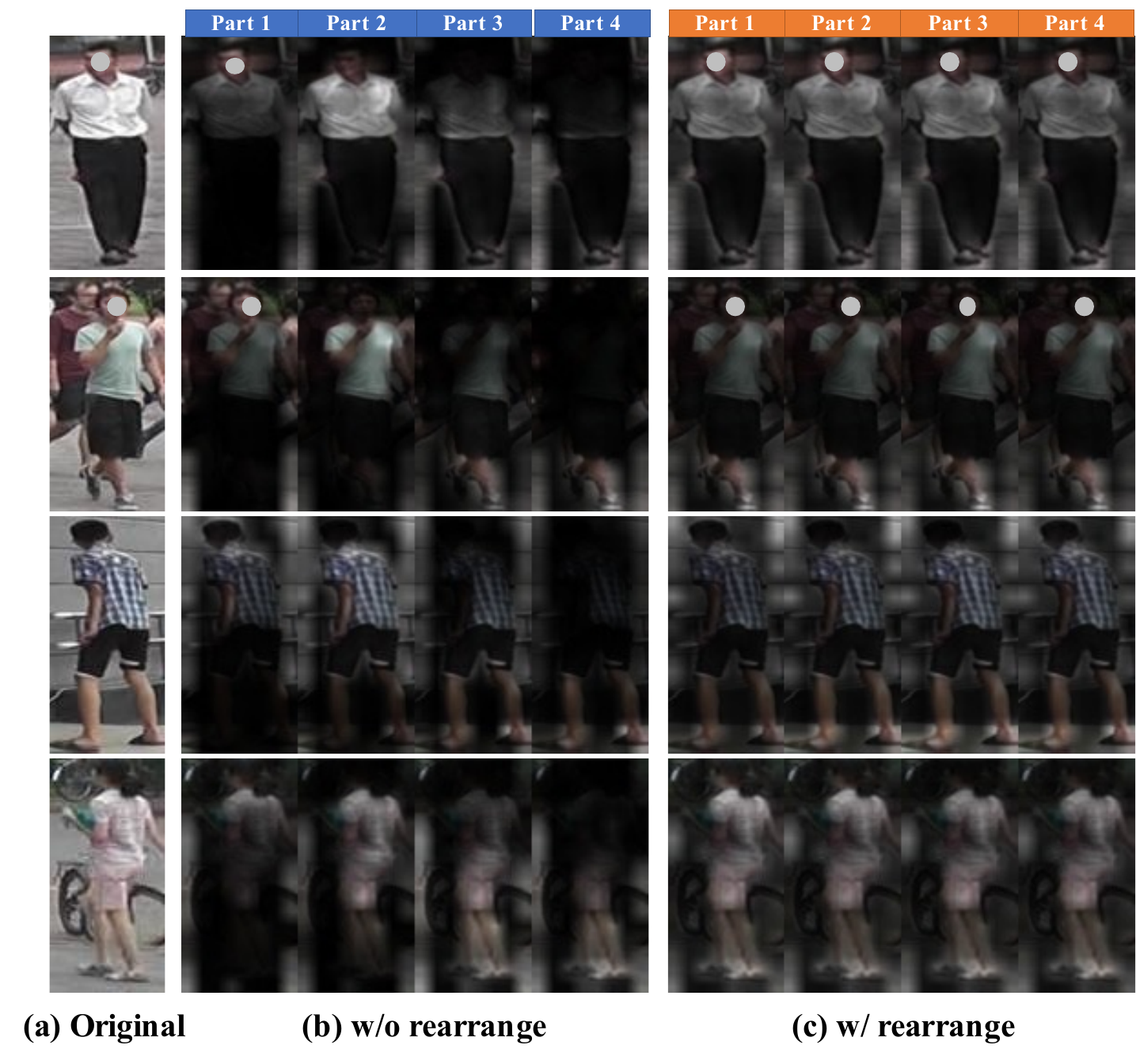}
	\vspace{-1.5em}
	\caption{Visualization of the learned attention masks for local features by JPM module. Higher weight results in higher brightness of the region. Note that we visualize the learned attention weights which are averaged among attention heads in the last layer. Faces in the images are masked for anonymization.} 
	\vspace{-1.em}
	\label{fig:vis}
\end{figure}

Although transformers can capture the global information in the image very well, a patch token still has a strong correlation with the corresponding patch. ViT-FRCNN \cite{ViT_FRCNN} shows that the output embeddings of the last layer can be reshaped as a spatial feature map that includes location information. In other words, if we directly divide the original patch embeddings into $k$ parts, each part 
may only consider a part of the continuous patch embeddings. Therefore, to better capture the long-range dependencies, we rearrange the patch embeddings and then re-group them into different parts, each of which contains several random patch embeddings of an entire image. In this way, the JPM module help to learn robust features with improved discrimination ability and more diversified coverage. 

To verify the above point, we visualize the learned attention of local features $[f_l^1,f_l^2,...,f_l^k]$ ($k=4$ in our cases) by JPM module in Figure~\ref{fig:vis}. Brighter region means higher corresponding weights. Several observations can be made from Figure~\ref{fig:vis}:
(1) The attention learned by the ``JPM w/o rearrange'' tends to focus on limited receptive fields (\ie the range of the corresponding patch sequences) due to global sequences being split into several isolated sub-sequences. For example, ``Part 1'' mainly pays attention to the head of a person, and ``Part 4'' is mainly focused around the bottom area. 
(2) In contrast, ``JPM w/ rearrange'' is able to capture long-range dependencies and each part has attention responses across the whole image because it is forced to extend its scope to the whole image through the rearranging operation. 
(3) According to the superior ReID performance and the intuitive visualization of rearranging effect, JPM is proved to not only capture more details at finer granularities but also learn robust and discriminative representations in the global context.

\section{More Visualization of Attention Maps}
\begin{figure*}[t]
    \centering
    \includegraphics[width=1.0\textwidth]{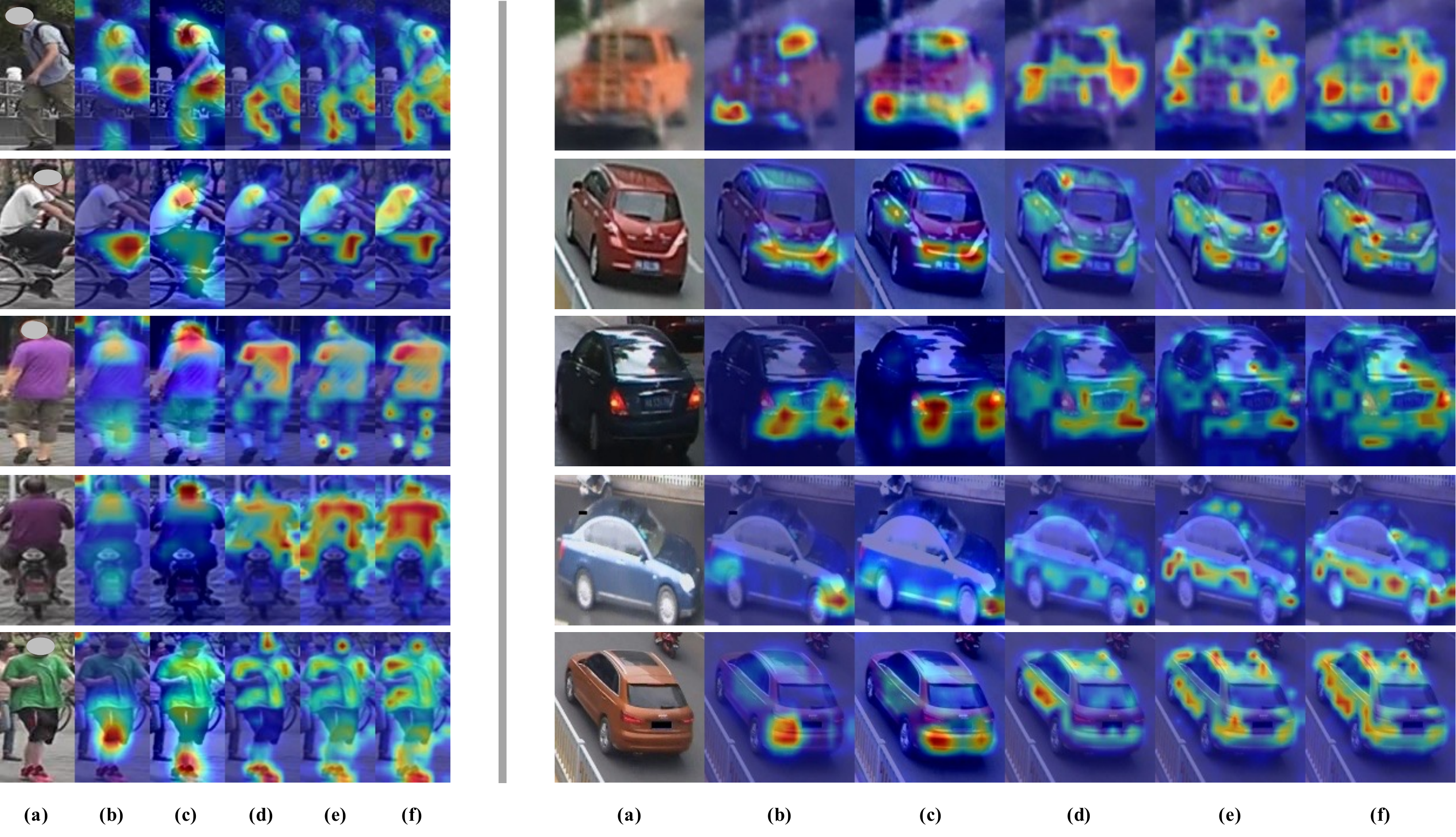}
	\vspace{-1.5em}
	\caption{Grad-CAM~\cite{grad_cam} visualization of attention maps. (a) Original images, (b) CNN-based methods, (c) CNN+Attention methods, (d) Transformer-based baseline, (e) TransReID w/o rearrange, (f) TransReID. Faces in the images are masked for anonymization.} 
	\vspace{-1.em}
	\label{fig:all}
\end{figure*}

In the main paper, we use Grad-CAM to visualize the gradient responses of our schemes, CNN-based methods, and CNN+attention methods. Following the similar setup, Figure~\ref{fig:all} shows more visualization results, with the similar conclusion that transformer-based methods capture global context information and more discriminative parts, which are further enhanced in our proposed TransReID for better performance.

\end{document}